\def\year{2016}
\documentclass[letterpaper]{article}
\usepackage{aaai}
\usepackage{times}
\usepackage{helvet}
\usepackage{courier}
\usepackage{epsfig}
\usepackage{graphicx}
\usepackage{float}	
\usepackage{tabularx}

\usepackage{multirow}
\usepackage{epstopdf}
\usepackage{pgfplots}
\usepackage{algorithmic}
\usepackage{algorithm}
\usepackage{subfigure}
\usepackage{amsmath}
\usepackage{amssymb}  

\begin{document}
%
\title{Improved Residual Vector Quantization for High-dimensional Approximate Nearest Neighbor Search}
\author{
  Liu Shicong, Lu Hongtao, Shao Junru\\
  \texttt{\{artheru, yz\_sjr, htlu\}@sjtu.edu.cn Shanghai Jiaotong University}
}
\maketitle
\begin{abstract}
\begin{quote}
Quantization methods have been introduced to perform large scale approximate nearest search tasks. 
Residual Vector Quantization (RVQ) is one of the effective quantization methods. 
RVQ uses a multi-stage codebook learning scheme to lower the quantization error stage by stage. 
However, there are two major limitations for RVQ when applied to on high-dimensional approximate nearest neighbor search: 
1. The performance gain diminishes quickly with added stages. 
2. Encoding a vector with RVQ is actually NP-hard. 
In this paper, we propose an improved residual vector quantization (IRVQ) method, 
our IRVQ  learns codebook with a hybrid method of subspace clustering and 
warm-started k-means on each stage to prevent performance gain from dropping, 
and uses a multi-path encoding scheme to encode a vector with lower distortion. 
Experimental results on the benchmark datasets show that our method gives substantially improves RVQ and delivers better performance compared to the state-of-the-art.

\end{quote}
\end{abstract}

\section{Introduction}
Nearest neighbor search is a fundamental problem in many computer vision applications 
such as image retrieval \cite{IR1} and image recognition \cite{IR2}. 
In high dimensional data-space, nearest neighbor search becomes very expensive 
due to the curse of dimensionality \cite{curse}. 
Approximate nearest neighbor (ANN) search is a much more practical approach. 
Quantization-based algorithms have recently been developed to perform ANN search tasks. 
They achieved superior performances against other ANN search methods \cite{pq}. 
Product Quantization \cite{pq} is a representative quantization algorithm. 
PQ splits the original $d$-dimensional data vector into $M$ disjoint sub-vectors 
and learn $M$ codebooks $\{\mathbf{C}_1\cdots\mathbf{C}_M\}$, 
where each codebook contains $K$ codewords $\mathbf{C}_m=\{\mathbf{c}_m(1),\cdots,\mathbf{c}_m(K) \}, m\in1\cdots M$. 
Then the original data vector is approximated by the Cartesian product of the codewords it has been assigned to. 
PQ allows fast distance computation between a quantized vector $\mathbf{x}$ and an input query vector $\mathbf{q}$ 
via asymmetric distance computation (ADC): the distances between $\mathbf{q}$ and all codewords $\mathbf{c}_m(k), 
m\in 1\cdots M, k\in 1\cdots K$ are precomputed, then the approximate distance between $\mathbf{q}$ and $\mathbf{x}$ 
can be efficiently computed by the sum of distances between $\mathbf{q}$ and codewords of $\mathbf{x}$ in $O(M)$ time. 
Compared to the exact distance computation taking $O(d)$ time, the time complexity is drastically reduced.

Product Quantization is based on the assumption that the sub-vectors are statistically mutual independent, 
such that the original vector can be effectively represented by the Cartesian product of quantized sub-vectors. 
However vectors in real data do not all meet that assumption. 
Optimized Product Quantization (OPQ) \cite{opq} and Cartesian K-means \cite{ck} 
are proposed to find an optimal subspace decomposition to overcome this issue.

Residual Vector Quantization (RVQ) \cite{rvq} is an alternative approach 
to perform approximate nearest neighbor search task. Similar to Additive Quantization (AQ) \cite{babenko2014additive} 
and Composite Quantization \cite{cq}, 
RVQ approximates the original vector as the sum of codewords instead of Cartesian product. 
Asymmetric distance computation can also be applied to data quantized by RVQ. 
RVQ adopts a multi-stage clustering scheme, on each stage the residual vectors 
are clustered instead of a segment of the original vector. 
Compared to PQ, RVQ naturally produces mutually independent codebooks. 
However, the gain of adding an additional stage drops quickly as residual vectors become more random, 
limiting the effectiveness of multi-stage methods to only a few stages \cite{gersho1992vector}. 
A direct observation is that the encodings of codebooks learned on the latter stages have low information entropy. 
Moreover, encoding a vector with dictionaries learned by RVQ is essentially a high-order Markov random field problem, which is NP-hard. 

In this paper, we propose the Improved Residual Vector Quantization (IRVQ). 
IRVQ uses a hybrid method of subspaces clustering and warm-started k-means to obtain high information entropy for each codebook, 
and uses a multi-path search method to obtain a better encoding. 
The basic idea behind IRVQ is rather simple:

\begin{enumerate}
\item Subspace clustering generally produces high information entropy codebook. 
Though we seek a clustering on the whole feature space, such codebook is still useful. 
We utilize these information by warm-start k-means with this codebook.

\item The norms of codewords reduce stage by stage. Though the naive "greedy" encoding fails to produce optimal encoding, 
a less "greedy" encoding is more likely to obtain the optimal encoding. 
We propose a multi-path encoding algorithm for learn codebooks.
\end{enumerate} 

The codebooks learned by IRVQ are mutually independent and each codebook has high information entropy. 
And a significantly lower quantization error observed compared to RVQ and other state-of-the-art methods. 
We have validated our method on two commonly used datasets for evaluating ANN search performance: 
SIFT-1M and GIST-1M \cite{pq}. The empirical results show that our IRVQ improves RVQ significantly. 
Our IRVQ also outperforms other state-of-the-art quantization methods such as PQ, OPQ, and AQ.

\section{Residual Vector Quantization}
Residual vector quantization (RVQ) \cite{juang1982multiple} is a common technique 
to approximate original data with several low complexity quantizers, 
instead of a prohibitive high complexity quantizer. 
RVQ reduces the quantization error by learning quantizers on the residues. RVQ is introduced to perform ANN-search in \cite{rvq},  The gain of adding an additional stage relies on the commonality among residual vectors from different cluster centers. Thus on high-dimensional data this approach performs badly.


\subsection{Information Entropy}

\begin{figure*}
\begin{center}
	\subfigure[Product Quantization]{
    	\includegraphics[width=0.22\linewidth]{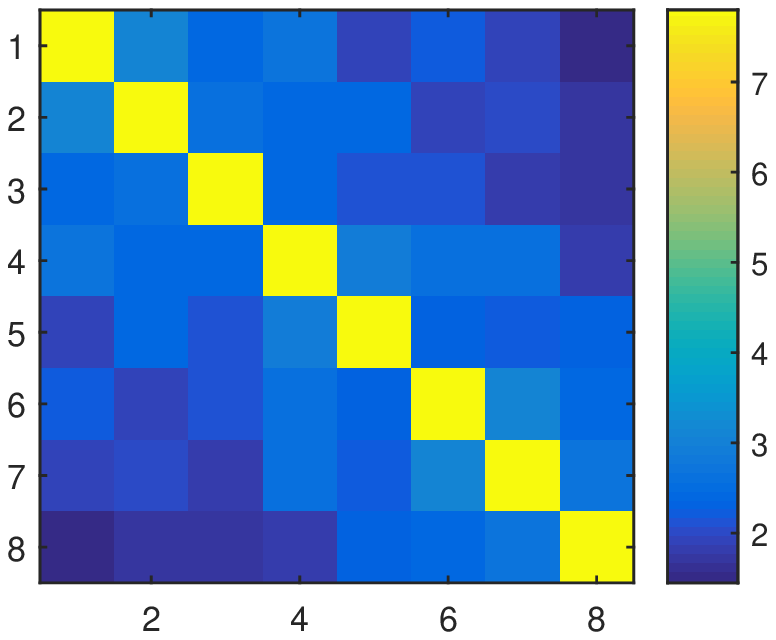}
    }
	\subfigure[Residual Vector Quantization]{
    	\includegraphics[width=0.22\linewidth]{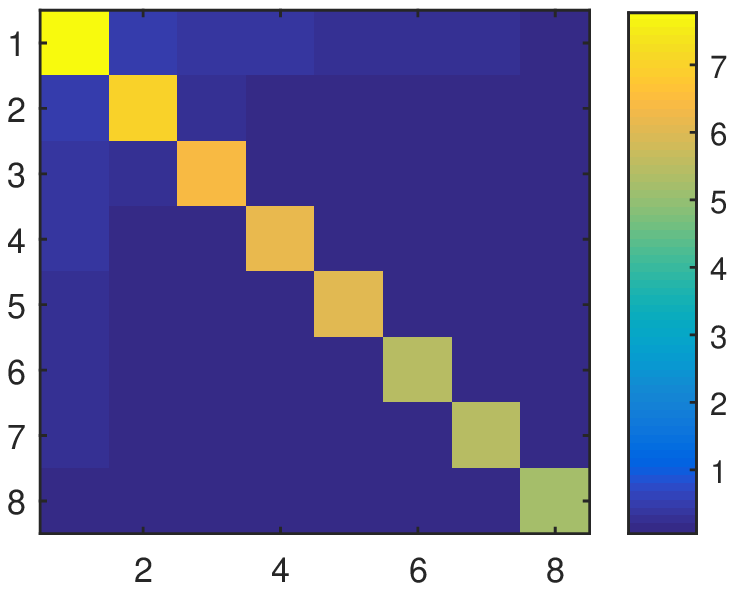}
    }
    	\subfigure[Improved Residual Vector Quantization($I=10, L=10$)]{
        	\includegraphics[width=0.22\linewidth]{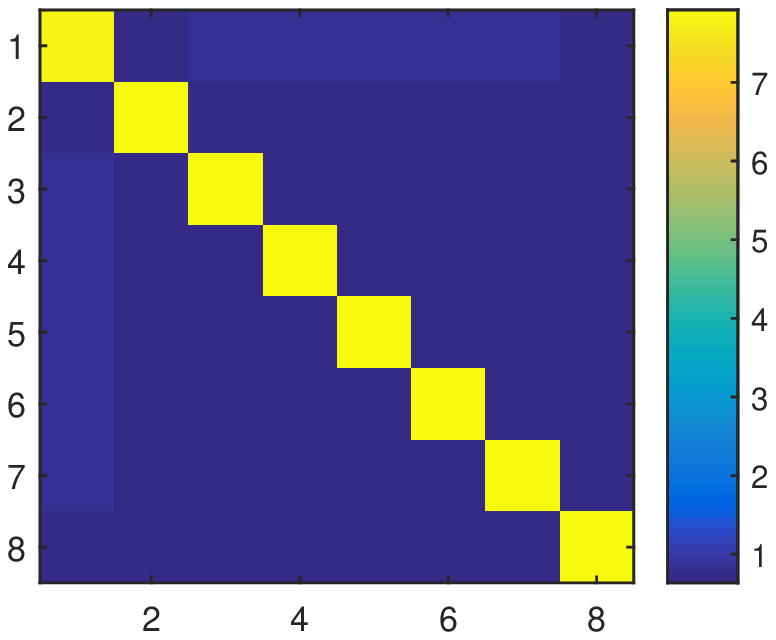}
        }           
        	\subfigure[Addtive Quantization]{
            	\includegraphics[width=0.22\linewidth]{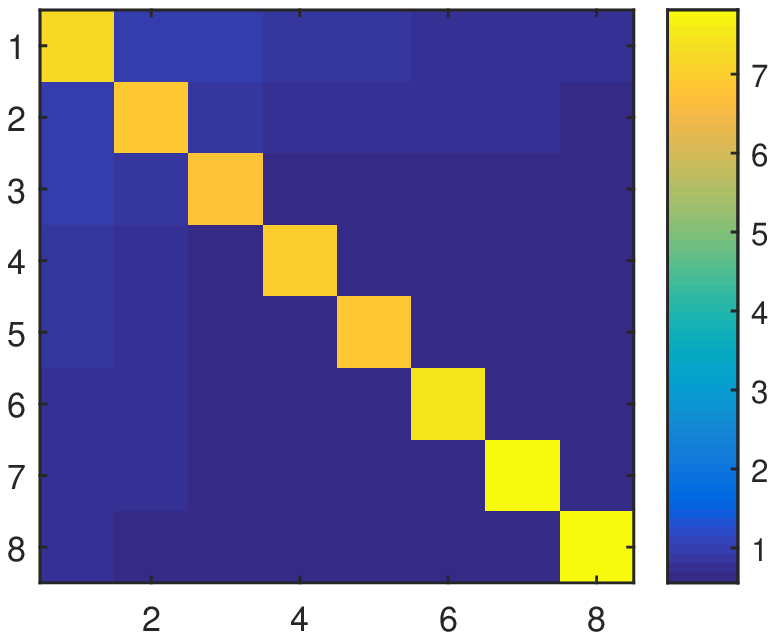}
            }           
\caption{Mutual information matrix for different quantization methods. Experiment conducted on a GIST-1M dataset. We learned $M=8$ codebooks, $K=256$ codewords per codebook. The perfect encoding should have no mutual information between different codebooks and an information entropy of $\log K=8$-bits for each codebook. Our proposed IRVQ achieves near optimal encoding.}
\label{figm}
  \end{center}
\end{figure*}

It has been observed that the residual vectors become very random with increasing stages, 
limiting the effectiveness of RVQ to a small number of stages. To begin with, 
we first examine the encoded dataset by RVQ from the point of view of information entropy.

\begin{figure}[t]
\begin{center}
   \pgfplotsset{grid style={dotted}}
   \pgfplotsset{every axis/.append style={font=\tiny, legend style={font=\small},}}
   	\begin{tikzpicture}
   		\begin{axis}[
   				title={SIFT1M},
   		width=0.25\textwidth,
   		height=0.2\textwidth,
   		axis y line=left,
   		axis x line=bottom,
   		xlabel=Stage,
   		xmin=0.5, xmax=16.5,
   		ymin=5,ymax=8,
   		xtick={2,4,6,8,10,12,14,16},
   		xlabel near ticks,
   		ylabel near ticks,
   		grid,
   		legend pos=south west,
   		legend to name=entroleg,
   		legend columns=-1,
   		legend entries={[black]RVQ,[black]{IRVQ($L$=1,$I$=10)},[black]{IRVQ($L$=10,$I$=10)}},
   		]
   		\addplot[color=blue, mark=triangle] table [x=m, y=plain, col sep=comma] {figures/entropySIFT.txt};
 		\addplot[color=red, mark=*] table [x=m, y=1-10, col sep=comma] {figures/entropySIFT.txt};
 		\addplot[color=black!30!green, mark=square] table [x=m, y=10-10, col sep=comma] {figures/entropySIFT.txt};
   		\draw[dashed](axis cs:0,0.3180) -- (axis cs:10,0.3180);
   		\node[anchor=west, fill=white] (source) at (axis cs:1,0.37){w/o quantization=0.3180};
   		\node (destination) at (axis cs:8,0.3180){};
   		\draw [->] (source)--(destination);
   		\end{axis}
   	
   	\end{tikzpicture}
 	   	\begin{tikzpicture}
 	   		\begin{axis}[
 	   				title={GIST1M},
 	   		width=0.25\textwidth,
 	   		height=0.2\textwidth,
 	   		axis y line=left,
 	   		axis x line=bottom,
 	   		xlabel=Stage,
 	   		xmin=0.5, xmax=16.5,
 	   		ymin=2,ymax=8,
 	   		xtick={2,4,6,8,10,12,14,16},
 	   		xlabel near ticks,
 	   		ylabel near ticks,
 	   		grid,
 	   		legend pos=south west
 	   		]
 	   		\addplot[color=blue, mark=triangle] table [x=m, y=plain, col sep=comma] {figures/entropyGIST.txt};
 	 		\addplot[color=red, mark=*] table [x=m, y=1-10, col sep=comma] {figures/entropyGIST.txt};
 	 		\addplot[color=black!30!green, mark=square] table [x=m, y=10-10, col sep=comma] {figures/entropyGIST.txt};
 	   		\draw[dashed](axis cs:0,0.3180) -- (axis cs:10,0.3180);
 	   		\node[anchor=west, fill=white] (source) at (axis cs:1,0.37){w/o quantization=0.3180};
 	   		\node (destination) at (axis cs:8,0.3180){};
 	   		\draw [->] (source)--(destination);
 	   		\end{axis}
 	   	
 	   	\end{tikzpicture}
 	   	\ref{entroleg}
\end{center}
   \caption{Entropy of each stage on SIFT1M and GIST1M datasets. Learned with $k=256$. For the codebook learned on stage $m$, information entropy $E=\sum_{n=1}^m p_n(\log_2 p_n)$, where $p_n=P(i_m(\mathbf{x})=n)$.}
      \label{figentropy}
\end{figure}
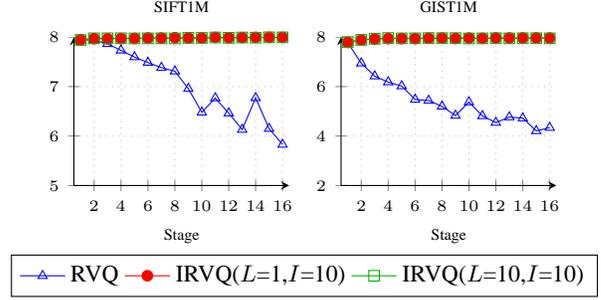

For hashing based approximate nearest neighbor search methods, e.g. Spectral Hashing \cite{SH}, 
we seek a code that each bit has a 50\% chance of being one or zero, and different bits are mutually independent. 
Similarly, we would like to obtain maximum information entropy $S(\mathbf{C}_m)$, defined below, 
for each codebook and no mutual information between different codebooks. 

\begin{equation}
\begin{split}
S(\mathbf{C}_m)=\sum_{k=1}^K p_k^m(\log_2 p_k^m)&=\log_2K \\
\sum_{k_i,k_j\in 1\cdots K} p_{ij}(k_i,k_j)&\log_2\frac{p_{ij}(k_i,k_j)}{p_{k_i}^i p_{k_j}^j}=0\\
& for\quad i,j\in 1\cdots M
\end{split}
\end{equation}

where $p_k^m$ denotes the probability that in the dictionary $\mathbf{C}_m$, 
its $k$-th element is chosen; and $p_{ij}(k_i,k_j)$ denotes the probability 
that $k_i$-th element from $\mathbf{C}_i$ and $k_j$-th element from $\mathbf{C}_j$ 
are chosen by a vector $\mathbf{x}$ simultaneously. 
In Fig. \ref{figentropy}, we drew the information entropy at each stage of 
the RVQ to reveal a phenomenon: For SIFT-1M dataset \cite{pq}, 
the entropy drops to 5.8bits at the 16th stage, and 4.34bits for GIST-1M. 
That means though we wish to learn $K$ codewords for each codebook, 
only a few of them are effective. 
Nevertheless, quantization on residual vectors leads to a much lower mutual information compared to PQ because residues are largely independent 
as shown in Fig.\ref{figm}.

\subsection{Encoding}
\label{secencoding}
RVQ encodes according to the distance between current residual vector and each codewords. 
We shall call it Sequential Encoding, which is actually a greedy algorithm. 
For codebooks learned with RVQ, 
The codewords chosen on a stage will affect the choice on the latter stages. 
Consider encoding a vector $\mathbf{x}$ with $M$ codebooks learned, the quantization error is:
\begin{equation}
\label{app}
\begin{split}
E &=\sum_{m=1}^{M'}\lVert \mathbf{x}-\mathbf{c}_m(i_m(\mathbf{x}))\rVert^2-(m-1)\lVert \mathbf{x}\rVert^2 \\
&\quad+\sum_{a=1}^{M'}\sum_{b=1,a\neq b}^{M'} \mathbf{c}_a(i_a(\mathbf{x}))^\mathrm{T}\mathbf{c}_b(i_b(\mathbf{x}))
\end{split}
\end{equation}

Denote $\epsilon$ as the third term. 
Sequential Encoding cannot predetermine the value of $\epsilon$. 
For PQ-based schemes such as Product Quantization \cite{pq}, 
Optimized Product Quantization \cite{opq}, 
such minimization scheme guarantees finding the best encoding, 
since codewords from different codebooks are mutually orthogonal. 
However, for RVQ, Sequential Encoding is not optimal. 
The encoding with codebooks learned by RVQ is actually a fully connected discrete pairwise MRF energy optimization, 
which is actually NP-hard. 

In addition, the quantization errors are propagated to the next stage in the codebook learning process. 
RVQ actually uses the information of the encoded vectors on its learning phase: 
the residual vectors are determined by both codebooks and the encoding mechanism. 
As RVQ for ANN search requires many stages of codebook learning, 
the accumulated quantization loss could become very high with Sequential Encoding.

\section{Improved Residual Vector Quantization}
As we have presented the two major drawbacks of RVQ. 
In this section, we propose our Improved Revised Residual Vector Quantization (IRVQ) to 
overcome these issues. 
IRVQ improves RVQ in three aspects:
\begin{description}
\item[Improved Codebook Learning] Instead of performing cold-started k-means as in RVQ, we warm-start k-means with initial codebook learned on a relatively smaller subspace.
\item[Multi-path Vector Encoding] In addition to assigning a vector to a single codeword, multiple codewords may be kept as candidates.
\item[Joint Optimization] On each stage, we first use Improved Codebook Learning to learn a codebook, then use Multi-path Vector Encoding to encode the dataset vectors, so the residual vectors could be even smaller.
\end{description} 

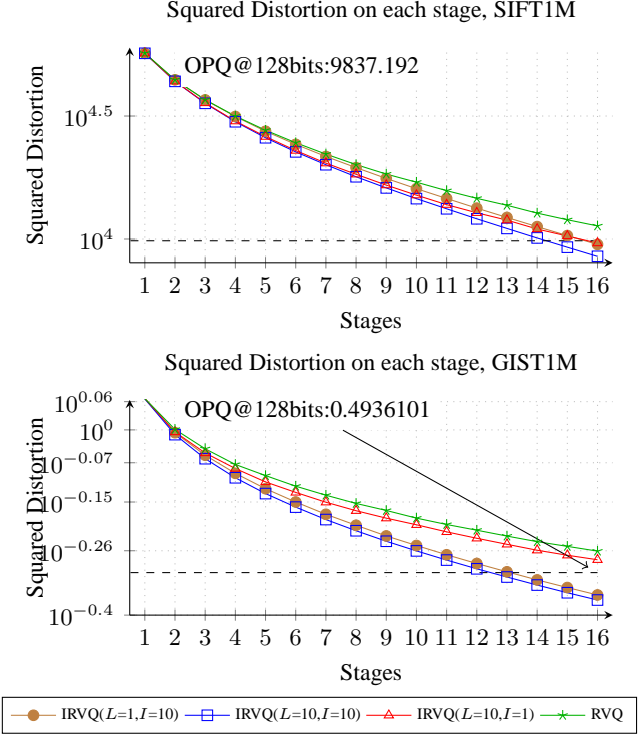
\begin{figure}[t]
\begin{center}
   \pgfplotsset{grid style={dotted}}
   \pgfplotsset{every axis/.append style={font=\small, legend style={font=\tiny},}}
   	\begin{tikzpicture}
   		\begin{axis}[
   					title={Squared Distortion on each stage, SIFT1M},
   			width=0.45\textwidth,
   			height=0.25\textwidth,
   			axis y line=left,
   			axis x line=bottom,
   			ylabel=Squared Distortion,
   			xlabel=Stages,
   			xtick={1,2,3,4,5,6,7,8,9,10,11,12,13,14,15,16},
   			xmin=0.5, xmax=16.5,
   			ymin=8000,ymax=60000,
   			ymode=log,
   			legend entries={[black]{IRVQ($L$=1,$I$=10)},[black]{IRVQ($L$=10,$I$=10)},[black]{IRVQ($L$=10,$I$=1)},[black]RVQ},
   			legend columns=-1,
   			legend to name=legS,
   			grid
   			]
   			\addplot[color=brown, mark=*] table [x=M, y=1-10, col sep=comma] {figures/shrinkSIFT.txt};
   			
   			\addplot[color=blue, mark=square] table [x=M, y=10-10, col sep=comma] {figures/shrinkSIFT.txt};

   			\addplot[color=red, mark=triangle] table [x=M, y=10-1, col sep=comma] {figures/shrinkSIFT.txt};
   			
   			\addplot[color=black!30!green, mark=star] table [x=M, y=plain, col sep=comma] {figures/shrinkSIFT.txt};

   			\draw[dashed](axis cs:0, 9837.192) -- (axis cs:16,9837.192);
   			\node[anchor=west, fill=white] (source) at (axis cs:2,50000){OPQ@128bits:9837.192};
   		\end{axis}
   	
   	\end{tikzpicture}
   		\begin{tikzpicture}
   			\begin{axis}[
   					title={Squared Distortion on each stage, GIST1M },
   			width=0.45\textwidth,
   			height=0.25\textwidth,
   			axis y line=left,
   			axis x line=bottom,
   			ylabel=Squared Distortion,
   			xlabel=Stages,
   			xtick={1,2,3,4,5,6,7,8,9,10,11,12,13,14,15,16},
   			xmin=0.5, xmax=16.5,
   			ymax=1.16, ymin=0.4,
   			ytick={1.15,1.0,0.85,0.7,0.55,0.4},
   			ymode=log,
   			grid
   			]
   			\addplot[color=brown, mark=*] table [x=M, y=1-10, col sep=comma] {figures/shrinkGIST.txt};
   			
   			\addplot[color=blue, mark=square] table [x=M, y=10-10, col sep=comma] {figures/shrinkGIST.txt};

   			\addplot[color=red, mark=triangle] table [x=M, y=10-1, col sep=comma] {figures/shrinkGIST.txt};
   			
   			\addplot[color=black!30!green, mark=star] table [x=M, y=plain, col sep=comma] {figures/shrinkGIST.txt};
   			   			
   			\draw[dashed](axis cs:0,0.4936101) -- (axis cs:16,0.4936101);
   			\node[anchor=west, fill=white] (source) at (axis cs:2,1.1){OPQ@128bits:0.4936101};
   			\node (destination) at (axis cs:16,0.4936101){};
   			\draw [->] (source)--(destination);
   			
   			\end{axis}
   		\end{tikzpicture}
   		   		\ref{legS}
\end{center}
   \caption{Distortion on SIFT1M and GIST1M, with $M=16$, $k=256$, also compared with Optimized Product Quantization. Note that $L$=10 $I$=1: use MVE only; $L$=1, $I$=10: use ICL only; $L$=10, $I$=10: Jointly use MVE and ICL}
\label{figShrink}
\end{figure}

See Algorithm \ref{algIOlearn} for the full pseudo code of IRVQ. 
The information entropy is maximized with our Improved Codebook Learning as shown in Figure \ref{figentropy}, 
and with Multi-path Vector Encoding better approximations of original vectors are found as shown in Figure \ref{figShrink}. 
We jointly use Improved Codebook Learning and Multi-path Vector Encoding to fully optimize the a stage, 
and the quantization error is even lowered. We shall explain our algorithm in more details in the following text.

\begin{algorithm}[t]

\caption{Improved Residual Vector Quantization}
\label{algIOlearn}
\textbf{Input}: Training samples $\mathbf{X}$ of $d$-dimension, number of stages $M$, number of centroids to learn on each stage $K$, best $L$ approximations on each stage, $I$ iterations.

\textbf{Output}: Codebooks: $\mathbf{C}_m=\{\mathbf{c}_m(1\cdots K), m=1\cdots M\}$.

\begin{algorithmic}[1]
\STATE Initialize Residue: $\mathbf{R}=\mathbf{X}$
\STATE Denote the best $L$ approximations for $\mathbf{x}$ on stage $m$: $\{\mathbf{x}^l_m, l=1\cdots L\}$
\FOR {$m=1\cdots M$}
\STATE Perform PCA on $\mathbf{R}$ and extract principal components  $\mathbf{A}=(\mathbf{r}_1,\mathbf{r}_2,\cdots,\mathbf{r}_d)$.
\STATE $\mathbf{C}'_m \leftarrow$ k-means on $(\mathbf{r}_1,\mathbf{r}_2,\cdots,\mathbf{r}_{\lceil d^{1/I}\rceil})^\mathrm{T}\mathbf{R}$
\FOR {$p=2\cdots I$}
\STATE $\mathbf{Y}\leftarrow(\mathbf{r}_1,\mathbf{r}_2,\cdots,\mathbf{r}_{\lceil d^{p/I}\rceil})^\mathrm{T}\mathbf{R}$
\STATE $\mathbf{C}'_m \leftarrow$ k-means on $\mathbf{Y}$: \\
		\quad Initialize k-means algorithm with $\mathbf{C}'_m$, missing dimensions are padded with zeros.
\ENDFOR
\STATE $\mathbf{C}_m=\mathbf{A}\mathbf{C}'_m$ is the codebook for stage $m$.
\FORALL {$\mathbf{x}\in\mathbf{X}$}
\STATE find the best $L$ approximations of $\mathbf{x} \approx \mathbf{x}_{m}^l=\mathbf{x}_{m-1}^{l'}+\mathbf{c}_m(k), l'=1\cdots L, k=1\cdots K$.
\ENDFOR
\STATE $\mathbf{r}=\mathbf{x}-\mathbf{x}_{m}^1$
\ENDFOR
\end{algorithmic}
\end{algorithm}

\subsection{Improved Codebook Learning (ICL)}
To obtain better clustering performance on high dimensional data, one of the popular approaches is to cluster on lower-dimensional subspace \cite{agrawal1998automatic}, this is also what PQ/OPQ do to obtain high information entropy for each dictionary. Many previously proposed methods for high dimensional data clustering, e.g. \cite{bouveyron2007high}, \cite{jing2007entropy}, seek clustering in an optimal subspace instead of the whole feature space. In lower-dimensional subspaces the projected datasets become denser and then a balanced clustering could be easily obtained. However, in the case of fitting the residual vectors, it's not reasonable to clustering on just a few dimensions as the residue vectors  lies in the whole feature space randomly. We thus seek a hybrid way to perform clustering on the residual vectors. 

We propose Improved Codebook Learning (ICL) to this end. It is done by iteratively adding dimensions for clustering. First designate a dimensions adding sequenc: $d_1 < d_2 < \cdots < d_I = d$, then, then:

\begin{description}
\item[Initialization] We first perform the principle component analysis (PCA) on the residual vectors $\mathbf{R}$ 
and extract the principal dimensions. 

\item[Iterative Warm-start] 
On the $i$-th iteration, perform k-means on top $d_i$ dimensions, initialized with previous learn codewords\footnote{On the first iteration we directly learn a codebook}. Note that the codewords have missing dimensions, we simply pad zeros to them.

\item[Restoration] Transform the codebook in PCA dimensions back to the original data-space dimensions. Then the resulting codebook can be used in the same way as the codebooks learned with usual k-means algorithm.

\end{description}

There are several motivations for our improved codebook learning method:
\begin{enumerate}
\item  As depicted in \cite{ding2004k}, PCA dimension reduction finds the best low-dimensional L2 approximation of the data.

\item Finding a better initial points for k-means almost always leads to a quicker 
convergence and a better clustering \cite{bradley1998refining}. 
As padding extra dimensions with zeros doesn't affect the encoding, 
it's reasonable to infer that the iterative initialization could also lead to better clustering.

\item PCA dimensions are essentially rotation of original data with no loss of information or metric change. Padding zeros doesn't change the clustering results.

\end{enumerate}
We warm-start k-means on the most significant PCA dimensions 
so clustering on the higher dimensions could started from better initial positions. In our experiment, a very low $I$ (in our experiments, $I$=10 at most), 
could already achieve satisfying results. By performing ICL, As shown in Figure \ref{figentropy}, 
the entropy of each codebook is maximized.

\subsection{Multi-path Vector Encoding (MVE)}
Encoding for Product Quantization is quite simple since the original feature space 
has been divided into mutually orthogonal subspaces. 
However, for Additive Quantization \cite{barnes1993vector}, Composite Quantization \cite{cq}, 
and Residual Vector Quantization, the encoding is an MRF problem as mentioned in Section \ref{secencoding}. Additive Quantization \cite{babenko2014additive} proposed a Beam Search algorithm in a matching pursuit fashion, however, runs slowly in practice\cite{babenko2015tree}.

\newcommand\Xtilde{\stackrel{\sim}{\smash{\mathbf{x}}\rule{0pt}{1.1ex}}}
Suppose the best approximation (correct encoding) of an input vector is 
$\mathbf{x}\approx\mathbf{c}_1(i_1)+\mathbf{c}_2(i_2)+\cdots+\mathbf{c}_M(i_M)$. 
Further assume we have known the first $m-1$ correct encodings ${i_1, i_2,\cdots, i_{m-1}}$, 
can we effectively compute $i_m$? 
Denote the known part as $\hat{\mathbf{x}}=\mathbf{c}_1(i_1)+\cdots+\mathbf{c}_{m-1}(i_{m-1})$  
and the unknown part as $\Xtilde=\mathbf{c}_{m+1}(i_{m+1})+\cdots+\mathbf{c}_{M}(i_{M})$, 
we seek the correct encoding on the $m$-th dictionary $i_m$. Notice that:
\begin{equation}
\begin{split}
\lVert\mathbf{x}-&\hat{\mathbf{x}}-\mathbf{c}_m(i_m)-\mathbf{x}'\rVert^2= \lVert\mathbf{x}-\hat{\mathbf{x}}\rVert^2 + \lVert\mathbf{x}-\Xtilde\rVert^2 + 2\hat{\mathbf{x}}^T\Xtilde\\
 & + \lVert\mathbf{x}-\mathbf{c}_m(i_m)\rVert^2 + 2\hat{\mathbf{x}}^T\mathbf{c}_m(i_m)+2\mathbf{c}_m(i_m))^T\Xtilde \\
 &-2\lVert\mathbf{x}\rVert^2
\end{split}
\end{equation}

The first three terms can be seen as constants when we seek the correct $i_m$ , 
and the last term can be omitted. The fourth and fifth term can be effectively computed. 
However the sixth term cannot be computed because we don't know $\Xtilde$. 
If we omit this term extra error will be introduced. To lessen this error, 
we hope $\lVert\Xtilde\rVert$ is very small so that the variance of 
the last term won't have an serious impact on the final outcome. 
The norms of codewords from codebooks learned with IRVQ naturally shrinks, 
so $\lVert\Xtilde\rVert$ is decreased stage by stage. 

Thus we maintain a list of best $L$ approximations of $\mathbf{x}$ with the first $(m-1)$ codebooks: 
$\{\mathbf{x}^1_{m-1},\mathbf{x}^2_{m-1}, \cdots , \mathbf{x}^l_{m-1}\}$. 
Then we encode with the next codebook $\mathbf{C}_m=\{\mathbf{c}_{m}(1), \mathbf{c}_{m}(2),\cdots,\mathbf{c}_{m}(K)\}$. 
We find $L$ combinations from $\{\mathbf{x}^{l}_{m-1}+\mathbf{c}_{m}(k)\}, l\in 1\cdots L, k\in 1\cdots K$ minimizing the following objective function: 
\begin{equation}
\begin{split} 
\lVert \mathbf{x}-\mathbf{x}^{l}_{m-1}-\mathbf{c}_{m}(k)\rVert^2=&\lVert\mathbf{x}-\mathbf{x}^{l}_{m}\rVert^2+\lVert\mathbf{x}-\mathbf{c}_{m}(k)\rVert^2 \\
&-\lVert \mathbf{x} \rVert^2 + 2\mathbf{c}_{m}(k)^T\mathbf{x}_{l}^{m-1}
\end{split}
\end{equation}

The first term has been computed at the previous encoding step, 
and the third term $\lVert \mathbf{x} \rVert^2$ is constant for any $(\mathbf{x}^{l}_{m-1}+\mathbf{c}_{m}(k))$, 
is thus negligible. And the last term involves $m$ table lookups and addition, 
with the inner-product of all codewords precomputed before MVE. 
Thus, only the term $\lVert\mathbf{x}-\mathbf{c}_{m}(k)\rVert^2$ is required to be computed. 
The time complexity is O($dK+mKL+KL\log L$) for encoding with one single dictionary on the $m$-th stage.

To sum up, MVE iteratively uses the top $L$ candidates as seeds to find the best encoding for $\mathbf{x}$. 
The resulting method is quite similar to the multi-path search for residual tree \cite{kossentini1992large}, 
which is used on signal reconstruction. Note that MVE degrades to Sequential Encoding when $L=1$. 

\subsection{Learning codebooks jointly with Improved Codebook Learning and Multi-path Encoding}
Remind that RVQ is a multiple stage learning procedure, at each stage the quantization error is left to the next iteration. 
This makes reducing quantization error for every single stage necessary. 
Now we have already presented the Improved Codebook Learning for learning balanced codewords, 
and Multi-path Vector Encoding for better encoding the original vector, 
we next jointly use them to optimize a stage in IRVQ. At each stage, our IRVQ does the following:
\begin{enumerate}
\item Use the Improved Codebook Learning to learn a codebook;
\item Use the Multi-path Vector Encoding to find a best approximation of the original vector with all the codebooks learned.
\end{enumerate}
Note that for each stage, the $L$ best approximations generated by MVE can be stored for future use, 
so the MVE on next stage only requires to encode one more codebook.

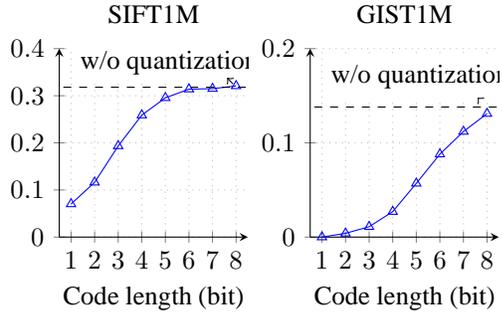
\begin{figure}
\begin{center}
   \pgfplotsset{grid style={dotted}}
   	\begin{tikzpicture}
   		\begin{axis}[
   				title={SIFT1M},
   		width=0.23\textwidth,
   		height=0.23\textwidth,
   		axis y line=left,
   		axis x line=bottom,
   		xlabel=Code length (bit),
   		xtick={1,2,3,4,5,6,7,8},
   		xmin=0.5, xmax=8.5,
   		ymin=0,ymax=0.4,
   		grid
   		]
   		\addplot[color=blue, mark=triangle] table [x=bits, y=SIFT1M, col sep=comma] {figures/epsilon.txt};
   		\draw[dashed](axis cs:0,0.3180) -- (axis cs:10,0.3180);
   		\node[anchor=west, fill=white] (source) at (axis cs:1,0.37){w/o quantization=0.3180};
   		\node (destination) at (axis cs:8,0.3180){};
   		\draw [->] (source)--(destination);
   		\end{axis}
   	
   	\end{tikzpicture}\begin{tikzpicture}
   			\begin{axis}[
   					title={ GIST1M},
   			width=0.23\textwidth,
   			height=0.23\textwidth,
   			axis y line=left,
   			axis x line=bottom,
   			xlabel=Code length (bit),
   			xtick={1,2,3,4,5,6,7,8},
   			ytick={0,0.1,0.2},
   			xmin=0.5, xmax=8.5,
   			ymin=0,ymax=0.2,
   			grid
   			]
   			\addplot[color=blue, mark=triangle] table [x=bits, y=GIST1M, col sep=comma] {figures/epsilon.txt};
   			\draw[dashed](axis cs:0,0.138) -- (axis cs:10,0.138);
   			\node[anchor=west, fill=white] (source) at (axis cs:1,0.17){w/o quantization=0.138};
   			\node (destination) at (axis cs:8,0.138){};
   			\draw [->] (source)--(destination);
   			\end{axis}
   		
   		\end{tikzpicture}
\end{center}
   \caption{How quantization on $\epsilon$ affects ANN search performance(Measured by Recall@1). Experimented on SIFT-1M and GIST-1M datasets, with $M=8, k=256, L=30, I=10$}
\label{figEpsilon}
\end{figure}
\subsection{Further Lower the Memory Consumption}
One of the advantage of quantization methods for ANN search is that vectors are compressed into 
a few bits so an in-memory search is feasible. Take PQ as an example, 
for an encoding with $M=8, K=256$, an original vector could be compressed into 
$M\log_2K=64$-bit code, thus a dataset containing 1 billion vectors can be fully loaded into a 8G RAM. 
For RVQ, AQ and our IRVQ, an extra storage overhead is required. 
Consider the expansion of distance between the approximated vector $\hat{\mathbf{x}}$ and the query vector $\mathbf{q}$:
\begin{equation}
\label{epsilonFix}
\begin{split}
 \lVert \mathbf{q}- \hat{\mathbf{x}} \rVert^2 & = \sum_{m=1}^M\lVert \mathbf{q}-\mathbf{c}_m(i_m(\mathbf{x})))\rVert^2-(m-1)\lVert \mathbf{q}\rVert^2 + \epsilon\\
\epsilon &= \sum_{a=1}^M\sum_{b=1,b\neq a}^M  {\mathbf{c}_a(i_a(\mathbf{x}))}^\mathrm{T}\mathbf{c}_b(i_b(\mathbf{x}))
\end{split}
\end{equation}

Note that $\epsilon$ is irrelevant to query $\mathbf{q}$, so we compute and store $\epsilon$ along with each quantized vector 
when the whole database is quantized, and this $\epsilon$ causes extra memory consumption. 
An IEEE-754 floating point number takes 32-bit of memory. 
However, it's possible to quantize $\epsilon$ into a few bits. 
Figure \ref{figEpsilon} shows that using only 8-bits of extra storage could already preserve the searching quality.

For real world ANN search applications, the major concern is the response speed instead of the memory consumption. 
Since an additional table look-up (indirect addressing) takes much more time compared to memory copy, 
if memory size is not an issue we recommend not to quantize $\epsilon$ for fastest speed.

\section{Approximate Nearest Neighbor Search Performance}
\label{test}
 \begin{figure*}
 \begin{center}
    \pgfplotsset{grid style={dotted}}
    \pgfplotscreateplotcyclelist{my}{
    	{blue, mark=*},
    	{teal, mark=o},
    	{yellow!40!red, mark=triangle},
    	{magenta, mark=diamond},
    	{black!20!green, mark=otimes},
    }
   	\begin{tikzpicture}
   		\begin{axis}[
   			title={Recall@$R$: SIFT1M, 64bit},
   			width=0.45\textwidth,
   			height=0.25\textwidth,
   			axis y line=left,
   			axis x line=bottom,
   			ylabel=recall@$R$,
   			xlabel=$R$,
   			xtick={1,2,4,8,16,32,64,128,256,512},
   			xticklabels={1,2,4,8,16,32,64,128,256,512},
   			ytick={0,0.1,0.2,0.3,0.4,0.5,0.6,0.7,0.8,0.9,1},   			
   			ymin=0.2, ymax=1,
   			xmode=log,
   			legend pos=south east,
   			grid,
   			cycle list name=my,
   			legend columns=-1,
   			legend entries={IRVQ, RVQ, PQ, OPQ, AQ},
   			legend to name=leg64
   			]
   			
   			
   			\addplot table [x=R, y=30-10, col sep=comma] {figures/recallSIFT1M64.txt};
   			

   			
   			   			
   			\addplot table [x=R, y=plain, col sep=comma] {figures/recallSIFT1M64.txt};
   			
   			\addplot table [x=R, y=PQ, col sep=comma] {figures/recallSIFT1M64.txt};

   			\addplot table [x=R, y=ckm, col sep=comma] {figures/recallSIFT1M64.txt};	

   			\addplot table [x=R, y=CQ, col sep=comma] {figures/recallSIFT1M64.txt};	
   			
   		\end{axis}
   	
   	\end{tikzpicture}
   	\begin{tikzpicture}
   		\begin{axis}[
   			title={Recall@$R$: GIST1M, 64bit},
   			width=0.45\textwidth,
   			height=0.25\textwidth,
   			axis y line=left,
   			axis x line=bottom,
   			ylabel=recall@$R$,
   			xlabel=$R$,
   			xtick={1,2,4,8,16,32,64,128,256,512},
   			xticklabels={1,2,4,8,16,32,64,128,256,512},
   			xmode=log,
   			ytick={0,0.1,0.2,0.3,0.4,0.5,0.6,0.7,0.8,0.9,1},   			
   			ymin=0.0, ymax=1,
   			legend pos=south east,
   			grid,
   			cycle list name=my,
   			]
   			
   			
   			\addplot table [x=R, y=30-10, col sep=comma] {figures/recallGIST1M64.txt};
   			

   			
   			   			
   			\addplot table [x=R, y=plain, col sep=comma] {figures/recallGIST1M64.txt};
   			
   			\addplot table [x=R, y=PQ, col sep=comma] {figures/recallGIST1M64.txt};

   			\addplot table [x=R, y=ckm, col sep=comma] {figures/recallGIST1M64.txt};	

   			\addplot table [x=R, y=CQ, col sep=comma] {figures/recallGIST1M64.txt};	
   			
   		\end{axis}
   		
   		\end{tikzpicture}
   		
    	\begin{tikzpicture}
    		\begin{axis}[
    			title={Recall@$R$: SIFT1M, 128bit},
    			width=0.45\textwidth,
    			height=0.25\textwidth,
    			axis y line=left,
    			axis x line=bottom,
    			ylabel=recall@$R$,
    			xlabel=$R$,
    			xtick={1,2,4,8,16,32,64,128},
    			xticklabels={1,2,4,8,16,32,64,128,256,512},
    			ytick={0,0.1,0.2,0.3,0.4,0.5,0.6,0.7,0.8,0.9,1},
    			xmin=1,xmax=128,
    			ymin=0.4, ymax=1,
    			xmode=log,
    			legend pos=north east,
    			grid,
    			cycle list name=my,
    			legend columns=-1,
    			legend entries={[black]IRVQ, [black]RVQ, [black]PQ, [black]OPQ, [black]AQ},
    			legend to name=leg128
    			]
    			
    			
    			\addplot table [x=R, y=30-10, col sep=comma] {figures/recallSIFT1M128.txt};
    			
 
    			
    			   			
    			\addplot table [x=R, y=plain, col sep=comma] {figures/recallSIFT1M128.txt};
    			
    			\addplot table [x=R, y=PQ, col sep=comma] {figures/recallSIFT1M128.txt};
 
    			\addplot table [x=R, y=ckm, col sep=comma] {figures/recallSIFT1M128.txt};	
 
    			\addplot table [x=R, y=CQ, col sep=comma] {figures/recallSIFT1M128.txt};	
    			
    		\end{axis}
    	
    	\end{tikzpicture}
    	\begin{tikzpicture}
    		\begin{axis}[
    			title={Recall@$R$: GIST1M, 128bit},
    			width=0.45\textwidth,
    			height=0.25\textwidth,
    			axis y line=left,
    			axis x line=bottom,
    			ylabel=recall@$R$,
    			xlabel=$R$,
    			xtick={1,2,4,8,16,32,64,128,256,512},
    			xticklabels={1,2,4,8,16,32,64,128,256,512},
    			xmode=log,
    			ytick={0,0.1,0.2,0.3,0.4,0.5,0.6,0.7,0.8,0.9,1},
    			ymin=0.1, ymax=1,
    			legend pos=south east,
    			grid,
    			cycle list name=my,
    			]
    			
    			
    			\addplot table [x=R, y=30-10, col sep=comma] {figures/recallGIST1M128.txt};
    			
 
    			
    			   			
    			\addplot table [x=R, y=plain, col sep=comma] {figures/recallGIST1M128.txt};
    			
    			\addplot table [x=R, y=PQ, col sep=comma] {figures/recallGIST1M128.txt};
 
    			\addplot table [x=R, y=ckm, col sep=comma] {figures/recallGIST1M128.txt};	
 
    			\addplot table [x=R, y=CQ, col sep=comma] {figures/recallGIST1M128.txt};	
    			
    		\end{axis}
    		
    		\end{tikzpicture}
    		\ref{leg128}
 \end{center}
    \caption{The performance for different algorithms on SIFT-1M and GIST-1M, with 128-bit encoding($M$=16)}
 \label{fig128b}
 \end{figure*}
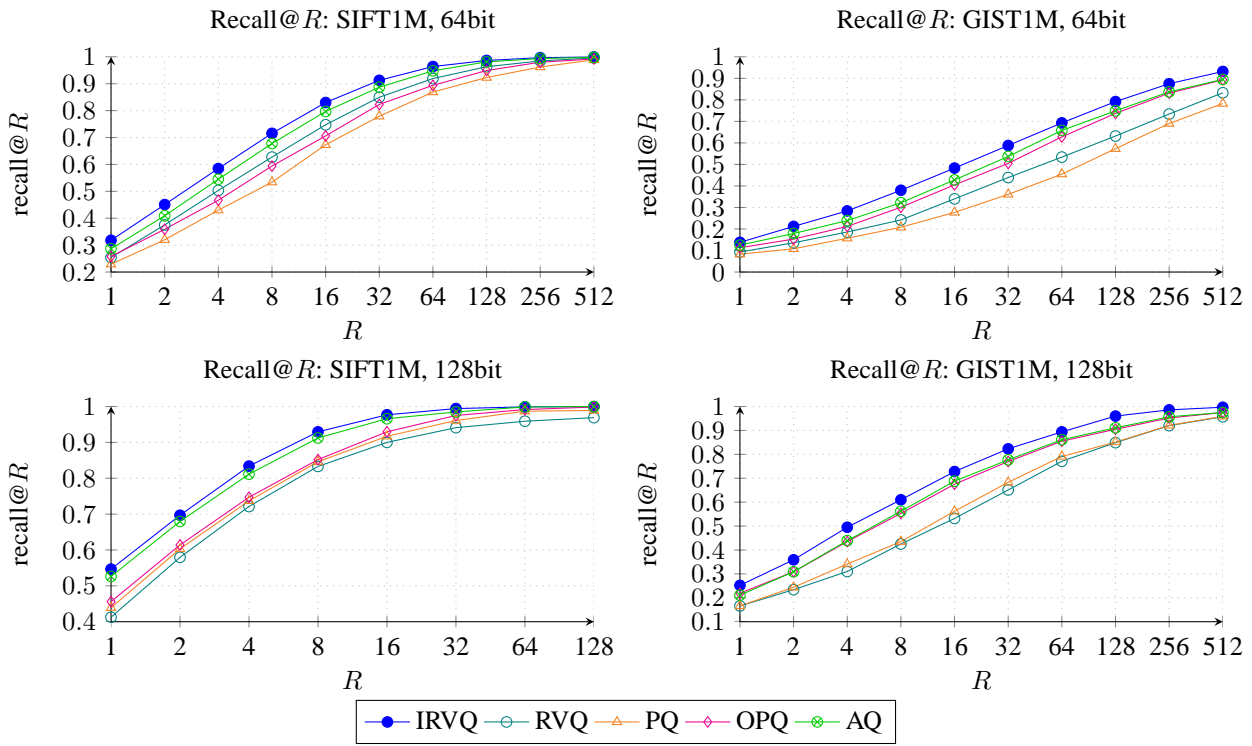
 
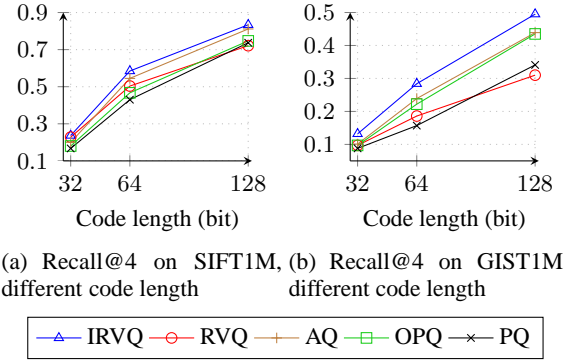
\begin{figure}
\begin{center}
   \pgfplotsset{grid style={dotted}}
       \pgfplotsset{every axis/.append style={font=\small, legend style={font=\small},}}
       \subfigure[Recall@4 on SIFT1M, different code length]{
   	\begin{tikzpicture}
   		\begin{axis}[
   		width=0.23\textwidth,
   		height=0.2\textwidth,
   		axis y line=left,
   		axis x line=bottom,
   		xlabel=Code length (bit),
   		xtick={32,64,128},
   		xmin=28,xmax=130,
   		ymin=0.1,ymax=0.9,
   		ytick={0.1,0.3,0.5,0.7,0.9},
   		grid
   		]
   		\addplot[color=blue, mark=triangle] table [x=bits, y=IO-30-10, col sep=comma] {figures/bitscompareSIFT.txt};
   		\addplot[color=red, mark=o] table [x=bits, y=plain, col sep=comma] {figures/bitscompareSIFT.txt};
   		\addplot[color=brown, mark=+] table [x=bits, y=cq, col sep=comma] {figures/bitscompareSIFT.txt};
   		\addplot[color=violet!20!green, mark=square] table [x=bits, y=ckm, col sep=comma] {figures/bitscompareSIFT.txt};
   		\addplot[color=black, mark=x] table [x=bits, y=pq, col sep=comma] {figures/bitscompareSIFT.txt};

   		\end{axis}
   	
   	\end{tikzpicture}
   	}
   	\subfigure[Recall@4 on GIST1M, different code length]{
   	\begin{tikzpicture}
   	\begin{axis}[
   		width=0.23\textwidth,
   		height=0.2\textwidth,
   		axis y line=left,
   		axis x line=bottom,
   		xlabel=Code length (bit),
   		xtick={32,64,128},
   		xmin=28,xmax=130,
   		ymin=0.05, ymax=0.5,
   		ytick={0.1,0.2,0.3,0.4,0.5},
   		grid,
   		legend columns=-1,
   		legend entries={[black]IRVQ, [black]RVQ, [black]AQ, [black]OPQ, [black]PQ},
   		legend to name=bitcmp
   		]
   		\addplot[color=blue, mark=triangle] table [x=bits, y=IO-30-10, col sep=comma] {figures/bitscompareGIST.txt};
   		\addplot[color=red, mark=o] table [x=bits, y=plain, col sep=comma] {figures/bitscompareGIST.txt};
   		\addplot[color=brown, mark=+] table [x=bits, y=cq, col sep=comma] {figures/bitscompareGIST.txt};
   		\addplot[color=violet!20!green, mark=square] table [x=bits, y=ckm, col sep=comma] {figures/bitscompareGIST.txt};
   		\addplot[color=black, mark=x] table [x=bits, y=pq, col sep=comma] {figures/bitscompareGIST.txt};   
   	\end{axis}
   		
   	\end{tikzpicture}
   	}
   	\ref{bitcmp}
\end{center}
   \caption{Effect of different code length for different ANN search methods. Experimented on datasets SIFT1M and GIST1M. The recall of the true neighbor in the top 4 ranked quantized element (Recall@4) is used to measure the ANN search quality.}
\label{figBits}
\end{figure}

We performed the ANN search tests on the two datasets commonly used to validate the efficiency of ANN methods: SIFT-1M and GIST-1M from \cite{pq}:
\begin{description}
\item[SIFT1M] contains one million of 128-d SIFT \cite{sift} features. It's commonly used with local feature descriptor for various image related applications.

\item[GIST1M] contains one million of 960-d GIST \cite{gist} global descriptors.
\end{description}
 
For each dataset, we randomly pick 100,000 vectors as the training set. We then encode the rest of the database vectors, and perform 1000 queries to check ANN search quality. 

\subsection{Evaluated Methods}
 
We compared our IRVQ to the following state-of-the-art quantization methods:
\begin{description}
\item[PQ]: Product quantization proposed in \cite{pq}. Following \cite{pq}, we used the structured ordering for GIST-1M and the natural ordering for SIFT-1M.
\item[OPQ]: Optimized Product Quantization proposed in \cite{opq}. We adopted the non-parametric version of OPQ. Cartesian k-means, the algorithm proposed in \cite{ck} shares a similar idea and has the same performance with OPQ.
\item[AQ]: Additive Quantization \cite{babenko2014additive}. We adopted the parameters suggest in the paper.
\item[RVQ]: Residual Vector Quantization proposed in \cite{rvq}.
\end{description}

For our IRVQ we set the parameters as $I=10, L=30$. For all the methods, we choose $K=256$ as the size of each codebook. We perform linear scan search with asymmetric distances computation (ADC) proposed in \cite{pq}, 
which directly compares the input query and the quantized dataset. 
The search quality is measured using recall@$R$, which means that for each query, 
we retrieved $R$ nearest items and check whether they contain the true nearest neighbor. 
Such criterion is commonly used to evaluate the ANN methods.

\subsection{Results}
Figure \ref{fig128b} shows the recall curve of various state-of-the-art methods. 
Our IRVQ improves RVQ significantly, for example, on 64bit encoding, IRVQ obtained \textbf{58.31\%} recall@4 for SIFT1M, 
while plain RVQ is only 50.35\%, the relative improvement is \textbf{15.8\%}. 
The improvement is even significant on higher dimensional data GIST1M, 
where IRVQ gained \textbf{28.4\%} and RVQ gained 18.6\% accuracy, relatively \textbf{52.7\%} improvement on recall@4. IRVQ also outperforms other state-of-the art, for example, IRVQ outperforms AQ by \textbf{17.7\%} on the recall@1 for 64bit SIFT1M encoding. 

We can also see by the results, RVQ doesn't perform well with added stages. On lower bits and low dimensions RVQ has advantages over PQ/OPQ, however, 
on higher bits or higher dimensions plain RVQ performs badly. 
Fig.\ref{figBits} illustrates the effect of code length for different methods. Our IRVQ fixed the problem and deliver  consistently high performance gain with added stages.

\section{Conclusion}
In this paper, we proposed the Improved Residue Vector Quantization (IRVQ) for large-scale high-dimensional approximate nearest neighbor search. We proposed Improved Codebook Learning (RCL) and the Multi-path Vector Encoding (MVE) to deliver consistent performance gain with adding stages. Experiment against several state-of-the-art quantization methods on two well known dataset demonstrate the effectiveness of IRVQ.

\bibliographystyle{aaai}
\bibliography{sigproc}  

\end{document}